\pdfoutput=1
\documentclass[11pt]{article}

\usepackage{acl}

\usepackage{times}
\usepackage{latexsym}

\usepackage[T1]{fontenc}
\usepackage[utf8]{inputenc}

\usepackage{microtype}

\usepackage{graphicx}
\usepackage{subfigure} 
\usepackage{amsmath}
\usepackage{amssymb}
\usepackage{threeparttable}
\usepackage{booktabs}
\usepackage{multirow}
\usepackage{makecell}
\usepackage{comment}
\DeclareMathOperator*{\argmax}{arg\,max}  
\usepackage[linesnumbered,ruled,vlined]{algorithm2e}
\SetCommentSty{mycommfont}
\SetKwInput{KwInput}{Input}                % Set the Input
\SetKwInput{KwOutput}{Output}              % set the Output

\makeatletter
\usepackage{enumitem}
\def\algbackskip{\hskip-\ALG@thistlm}
\makeatother
\usepackage{pifont}
\usepackage{xcolor}
\usepackage{tabularx}

\usepackage{threeparttable}
\usepackage{booktabs}
\usepackage{multirow}
\usepackage[
  separate-uncertainty = true,
  multi-part-units = repeat
]{siunitx}
\usepackage{graphicx, subcaption}
\usepackage{tikz}

\usepackage{array}

\newcolumntype{B}{!{\hspace{-1.5ex}}c}
\newcolumntype{D}{!{\hspace{-2ex}}c}
\newcolumntype{A}{!{\hspace{-1.2ex}}l}

\title{DEGAP: Dual Event-Guided Adaptive Prefixes for Templated-Based Event Argument Extraction with Slot Querying}

\author{
    Guanghui Wang$^{1}$, Dexi Liu$^{1*}$, Jian-Yun Nie$^{2}$, Qizhi Wan$^{1}$, \\ \textbf{Rong Hu$^{1}$, Xiping Liu$^{1}$, Wanlong Liu$^{3}$ and Jiaming Liu$^{4}$}  \\
    $^{1}$School of Information Management, Jiangxi University of Finance and Economics \\
    $^{2}$University of Montreal (UdeM), Department of Computer Science and Operations Research (DIRO) \\ 
    $^{3}$School of Computer Science and Engineering, University of Electronic Science and Technology of China \\
    $^{4}$Department of Statistics, Division of the Physical Sciences, the University of Chicago \\
    \texttt{\{15279836691, dexi.liu\}@163.com}
}

\begin{document}
\maketitle

\begin{abstract}
Recent advancements in event argument extraction (EAE) involve incorporating useful auxiliary information into models during training and inference, such as retrieved instances and event templates. These methods face two challenges: (1) the retrieval results may be irrelevant %insufficient utilization of relevant event instances due to deficiencies in retrieval 
and (2) templates are developed independently for each event without considering their possible relationship. %missing important information provided by relevant event templates. 
In this work, we propose DEGAP to address these challenges through a simple yet effective components: dual prefixes, i.e. learnable prompt vectors, where the instance-oriented prefix and template-oriented prefix are trained to learn information from different event instances and templates. Additionally, we propose an event-guided adaptive gating mechanism, which can adaptively leverage possible connections between different events and thus capture relevant information from the prefix. Finally, these event-guided prefixes provide relevant information as cues to EAE model without retrieval. Extensive experiments demonstrate that our method achieves new state-of-the-art performance on four datasets (ACE05, RAMS, WIKIEVENTS, and MLEE). Further analysis shows the impact of different components. %the proposed design and the effectiveness of the main components.
\end{abstract}

\section{Introduction}
\label{introduction}
\begin{figure}[t]
    \centering
    \includegraphics[width=\linewidth]{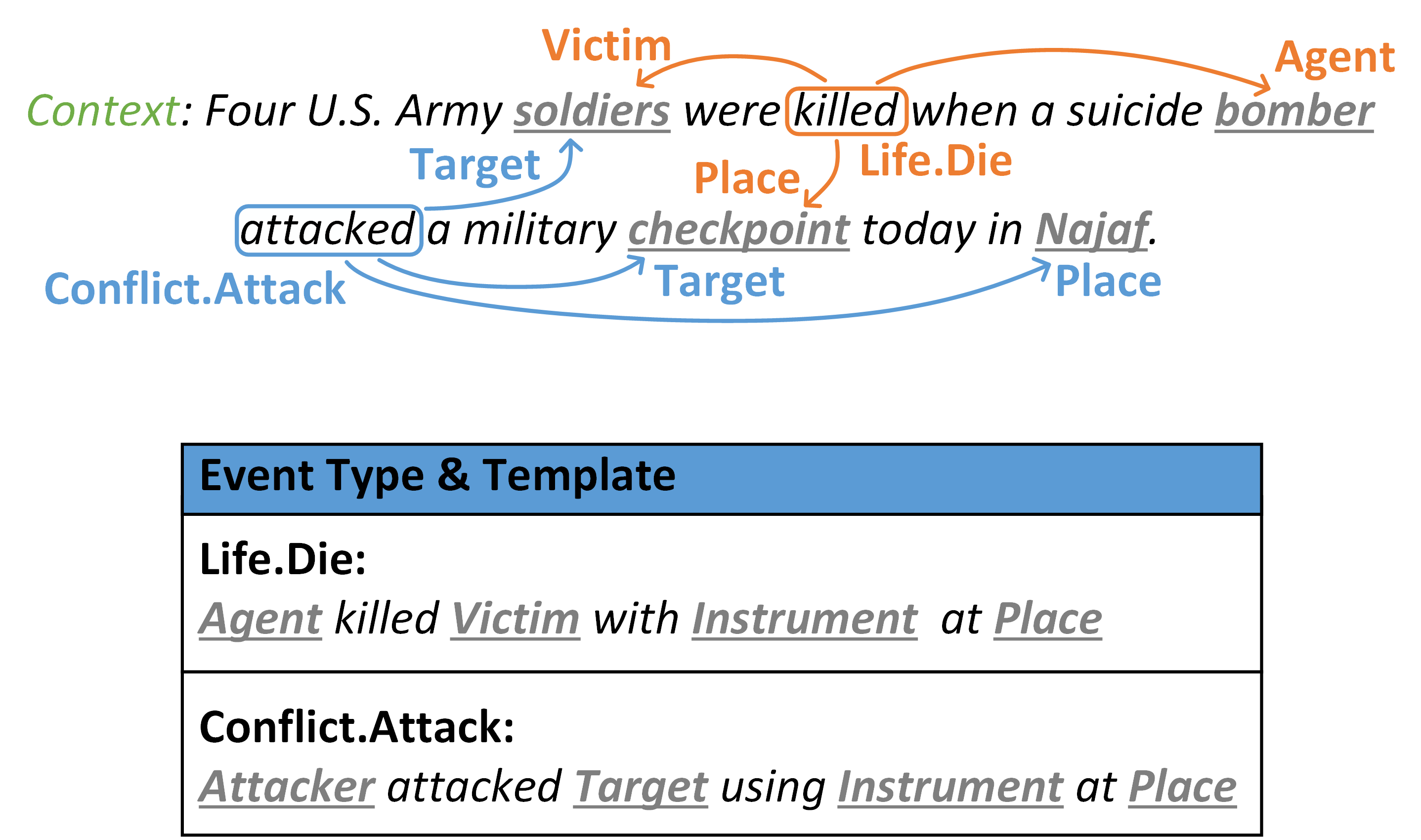}
    \caption{Two examples of EAE and event template from \citet{ma2022prompt}. Each underlined part (i.e., role slot) in the template is named after an argument role.}
    \label{fig1}
\end{figure}

As an essential subtask of Event Extraction, Event Argument Extraction (EAE) aims to identify event arguments and their roles for given event mentions in context. For example, in Fig.~\ref{fig1}, given a \textit{Life.Die} event with trigger \textit{killed}, EAE models need to extract arguments \textit{soldiers}, \textit{bomber}, and \textit{checkpoint}, and the corresponding roles --- \textit{Victim}, \textit{Agent}, and \textit{Place}. 
In the field of NLP, EAE has always been a challenging task, which requires deep natural language understanding \cite{hsu2023ampere}.

Consequently, some efforts incorporate ancillary information useful for EAE into models during training and inference, such as augmenting input with retrieved instances \cite{du-ji-2022-retrieval, ren-etal-2023-retrieve, zhao2023demosg} or querying arguments using role slots in event template \cite{liu2024beyond,  ma2022prompt, he2023revisiting, nguyen2023contextualized}. Moreover, in light of advancements in prompt learning \cite{li-liang-2021-prefix, liu-etal-2022-p}, some studies introduce prefixes (i.e, learnable prompt vectors) into models as additional keys and values to influence the computation of attention \cite{liu2022dynamic, zhang-etal-2023-overlap, hsu2023ampere, nguyen2023contextualized}. Despite the empirical success of previous studies, two issues remain, % that have not been appropriately solved, 
hindering further performance improvements.

\textbf{First, retrieved instances are not always reliable due to deficiencies in retrieval}. Previous studies regard event instances retrieved from training data \cite{wang2022training} as cues to extract arguments.
% and learn by analogy
However, these methods typically use a retriever pre-trained on other tasks with frozen parameters, which may not always retrieve %is often insufficient in recalling 
other event instances relevant to the target event \footnote{In a context, only a few words are event arguments, while other distracting context can mislead retrieval based on sentence semantic similarity. 
According to \cite{ren-etal-2023-retrieve}, on the RAMS dataset \cite{ebner-etal-2020-multi}, only 16.51\% of event instances can recall a sample with the same event type.
}.
Additionally, these methods often suffer from high computational costs and nontrivial knowledge loss because they independently retrieve and encode event instances, resulting in a poor local optimum. Directly increasing the number of instances to be retrieved is typically inefficient. 
Recent advancements show that one can \textit{utilize prefixes as efficient parameters to learn information from different corpora and integrate retrieved instances into prefixes as cues for inference} \cite{zhang2023overlap, liu-etal-2023-recap}. Inspired by this intriguing observation, in this work, we attempt to explore a novel approach to utilize relevant event instance without retrieval. First, we use prefixes to learn the semantic information of event instances. Then, we leverage the learned prefixes as cues to guide the model.

\textbf{Second, no relation between different %other relevant 
event templates is considered.} Most of the existing approaches  use a separate template as prompt for each target event. However, some event templates for other relevant events may provide useful information to the given event. This potential connection between event templates has been overlooked in the literature \cite{ma2022prompt, he2023revisiting}. Only \cite{ren-etal-2023-retrieve} attempted to retrieve role labels from other events that are relevant to the target event as demonstrations. % and achieved success. 
As shown in Fig.~\ref{fig1}, event templates provide rich semantic information about events, and other relevant events often have similar structures and share part of their roles (e.g., \textit{Life.Die} and \textit{Conflict.Attack}) \cite{li-etal-2023-intra}. Intuitively, one can regard relevant event templates as cues and perform reasoning by analogy. This can help EAE models better extract the target event from complex semantics.

In this paper, we aim to address the above challenges. We propose DEGAP (\textbf{D}ual \textbf{E}vent-\textbf{G}uided \textbf{A}daptive \textbf{P}refixes for event argument extraction). Specifically, we design two types of prefixes: the instance-oriented prefix and the template-oriented prefix. The reason we design separate prefixes for
event instances and templates, rather than a shared
prefix for both, is that instances and templates contain different types of information about events: an instance is a concrete example of an event, while a template tries to capture the general structure of an event. They play different roles in EAE, thus it is more appropriate to train separate prefixes for them\footnote{Our experimental results also confirm that dual prefixes can better learn semantic information from different event instances and templates, and further guide model to extract argument (\S~\ref{Why are Dual Prefixes Necessary?}).}.
These instance/template-specific parameters are used to learn information from different event instances/templates and enable them to interact sufficiently with each other, rather than ignoring the potential connections between them.
They are trained and updated together with model parameters, thus avoiding the insufficient retrieval and poor local optimum issues present in retrieval-based EAE methods. 
Moreover, not all event instances and templates are useful to the target event. We aim to use only those instances and templates that are relevant to the target event \cite{xu-etal-2022-two,liu2023enhancing}.
To connect other relevant
events to the target event and thus adaptively capture relevant information from the prefix, we design an event-guided adaptive gating mechanism. 
It utilizes the target event's trigger and type to score each instance-oriented and template-oriented prefix token via gated weight assignment at each layer, respectively. 
Finally, these event-guided prefixes are used to guide model for extracting argument.

To comprehensively evaluate our method, we conduct extensive experiments on four benchmark datasets (ACE05, RAMS, WIKIEVENTS, MLEE). Experimental results demonstrate that our method outperforms existing methods across all datasets, illustrating the effectiveness of our approach. Furthermore, we conduct a thorough study to explore the reasons for the performance improvement of our method, confirming the significance of the proposed design and the efficacy of main components.

\section{Related Work}
% \textbf{Event argument extraction}. Traditionally, most models are classification-based \cite{chen-etal-2015-event, nguyen-etal-2016-joint-event, yang-etal-2018-dcfee, ebner-etal-2020-multi, zhang-etal-2020-two, du-cardie-2020-document, xu-etal-2022-two}. Recently, there has been a trend towards framing EAE as a text generation or question-answering (QA) task. Generation-based methods sequentially generate all arguments of the target event by following a pre-defined template \cite{li-etal-2021-document, hsu2021degree, zeng-etal-2022-ea2e, hsu2023ampere}. QA-based methods query the arguments of each role using corresponding question templates \cite{du-cardie-2020-event, liu-etal-2021-machine, wei-etal-2021-trigger, lu-etal-2023-event, uddin2024generating} or role slots in event template \cite{zhang-etal-2022-transfer, ma2022prompt, li-etal-2023-intra, he2023revisiting}.

% This work builds on the existing state-of-the-art (SOTA) methods PAIE \cite{ma2022prompt} and TabEAE \cite{he2023revisiting}, which solves EAE by using role slots in the event template to query input context for extracting argument spans. With two proposed simple yet effective core components—the dual prefixes and the event-guided adaptive gating mechanism—our method establishes new SOTA performance on four datasets.

% \vspace{3pt}
% \noindent
\textbf{Event (argument) extraction with auxiliary information}. As the task requires a deep understanding of context, many previous efforts have been put into exploring which auxiliary information is useful for event predictions. 
One line of studies is incorporating such auxiliary information during training and inference, such as event templates \cite{li-etal-2021-document, hsu2021degree, zeng-etal-2022-ea2e, he2023revisiting, liu2024beyond}, natural questions \cite{liu2020event, liu-etal-2021-machine, lu-etal-2023-event, uddin2024generating}, retrieved instances \cite{du-ji-2022-retrieval, ren-etal-2023-retrieve, zhao2023demosg}, 
already predicted events \cite{huang-etal-2023-simple, du-etal-2022-dynamic}, 
and syntactic and semantic information \cite{huang2017zero,veyseh2020graph,ahmad2021gate,xu-etal-2022-two,yang2023amr}. 
As mentioned in \S~\ref{introduction}, these methods may lead to irrelevant results %1) fail to fully utilize relevant event instances 
due to retrieval deficiencies that may negatively impact the final results. Our approach is based on the principle of prompt learning. 

\vspace{5pt}
\noindent
\textbf{Prompt learning}. Prompt learning is a straightforward and effective method that adjusts PLMs for specific downstream tasks by adding prompts to the input. One line of research applies an automated search to find appropriate discrete prompting words \cite{wallace-etal-2019-universal, shin-etal-2020-autoprompt, wen2024hard}. An alternative line of research attaches a set of trainable prefix tokens to the model's attention module as additional keys and values \cite{li-liang-2021-prefix, liu-etal-2022-p,chen2023one}. This method has been widely applied in various studies. For example, \citet{liu-etal-2023-recap} integrates retrieved instances into the prefix as cues for inference. \citet{zhang2023overlap} uses prefixes to learn information from different corpora and refine inference. 
Additionally, some EAE efforts use prefixes to optimize manually designed prompts \cite{liu2022dynamic, nguyen2023contextualized} or to integrate syntactic and semantic information \cite{hsu2023ampere}.

Inspired by these advancements \cite{zhang2023overlap, liu-etal-2023-recap}, We explore a novel approach to utilize potentially relevant event information rather than retrieval: using prefix to learn the semantic information of different event instances and templates, then leveraging the learned prefix as cues to guide model. 
Moreover, previous methods overlook the potential connections between event templates. Some relevant event templates may provide useful information to the target event.
As such, we propose an event-guided adaptive gating mechanism, which can effectively connect other relevant events to the target event and thus adaptively capture relevant information from the prefix.

\section{Methodology}
In this section, we first define the EAE task. Then, we detail how the proposed DEGAP solves the challenges mentioned.

% \vspace{pt}
\noindent
\textbf{Task Definition} Formally, we first define a set of event types $\mathcal{E} = \{e_1, e_2, e_3, \ldots\}$ and a corresponding set of roles $R^e = \{r_1,r_2,r_3,\ldots\}$ for each event type $e \in \mathcal{E}$. Then, given a context $X=\{x_1,x_2,\ldots,x_n\}$ of $n$ words, the event trigger $x_{\text{trig}} \in X$ and the event type $e$, the task aims to extract all $(s,r)$ pairs for the event, where $s$ is a text span in $X$, and $r$ is an argument role in $R^e$.

\subsection{Model Architecture}

\begin{figure*}[t]
  \centering
  \includegraphics[width=\textwidth]{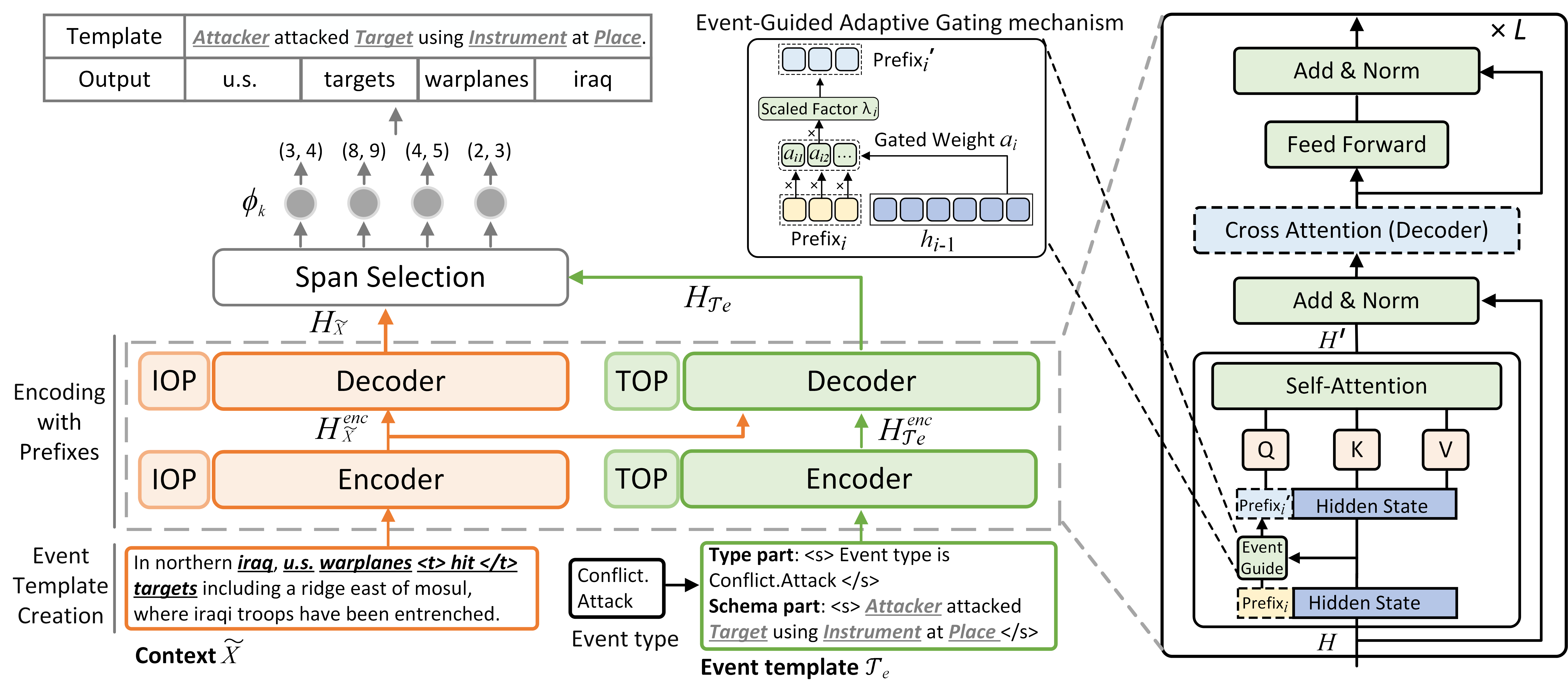}
  \caption{The overview of DEGAP. Here, the two encoders share the same parameters, and the two decoders also share the same parameters.
  % Additionally, $\text{P}_\text{k}$ and $\text{P}_\text{v}$ share the same prefix token parameters. 
  Due to space constraints, the event template here is simplified. When the model input is context, \( \text{Prefix}_i \) and \( h_{i-1} \) refer to \( \text{Prefix}_{i}^{\textit{ins}} \) and \( h_{i-1}^{trig} \), respectively. Conversely, they refer to \( \text{Prefix}_{i}^{\textit{tem}} \) and \( h_{i-1}^{type} \), respectively.  
  % When decoding event template. $H_{\widetilde{X}}^{enc}$ is fed into the decoder for computing a context-aware template representation through cross-attention mechanism. When decoding the context, the cross-attention mechanism is skipped.
  \textbf{IOP}: Instance-Oriented Prefix.
 \textbf{TOP}: Template-Oriented Prefix. 
 }
  \label{fig2}
\end{figure*}

Fig.~\ref{fig2}  shows the overall architecture of DEGAP. For a given event type $e$, we create an event template for it (\S~\ref{Event Template Design}). To learn information from event instances and templates, we design instance-oriented prefixes and template-oriented prefixes, respectively (\S~\ref{dual prefixes}). Dual prefixes can better learn semantic information from different event instances and templates compared to a shared prefix (\S~\ref{Why are Dual Prefixes Necessary?}). Then, we design an event-guided adaptive gating mechanism to connect other relevant
events to the target event and thus adaptively capture relevant information in the prefix (\S~\ref{Event-Guided Adaptive Gating mechanism}). Subsequently, these event-guided prefixes provide relevant event information as cues to model (\S~\ref{Encoding with Prefixes}). Finally, we introduce the span selection, which utilizes role slots in the event template to query context for extracting argument (\S~\ref{Span Selection}).

\subsubsection{Event Template Creation}
\label{Event Template Design}
We create an event template $\mathcal{T}_{e}$ for the given event type $e$, utilizing its rich semantic information to guide the model \cite{hsu-etal-2022-degree}. Similar to previous works \cite{liu2022dynamic, zhang2023overlap}, the event template $\mathcal{T}_{e}$ consists of two parts: the type part $\mathcal{I}_{e}$ and the schema part $\mathcal{S}_{e}$. The type part $\mathcal{I}_{e}$ adopts the format ``Event type is \texttt{[Event type]}”. For the schema part $\mathcal{S}_{e}$, we directly use that from TabEAE \cite{he2023revisiting}. 
Considering the example in Fig.~\ref{fig2}, where there is a \textit{Conflict.Attack} event with the event trigger \textit{hit}, the template for this event is: “\texttt{<s>} Event type is Conflict.Attack \texttt{</s>} \texttt{<s>} \underline{\textit{Attacker}} (and \underline{\textit{Attacker}}) attacked \underline{\textit{Target}} (and \underline{\textit{Target}}) using \underline{\textit{Instrument}} (and \underline{\textit{Instrument}}) at \underline{\textit{Place}} (and \underline{\textit{Place}}) \texttt{</s>}”, where each underlined part (i.e., role slot) is named after an argument role. The brackets indicate that there may be multiple arguments of that role in an event.

\subsubsection{Dual Prefixes}
\label{dual prefixes}
To directly and fully benefit from relevant event instances and templates without %regarding 
retrieval as an intermediate step, we construct two types of prefixes: (1) the \textbf{I}nstance-\textbf{O}riented \textbf{P}refix (IOP) and (2) the \textbf{T}emplate-\textbf{O}riented \textbf{P}refix (TOP). 
These instance/template-specific prefix parameters can learn rich semantic information from different event instances/templates and are trained and updated alongside model parameters. 

Note that we only design a shared IOP/TOP for all event instances/templates, rather than one for each event instance/template. This approach allows them to interact fully with each other, rather than ignoring the potential connections between them.
% thereby avoiding insufficient retrieval and poor local optimum. 
% Subsequently, these learned prefixes guide model to extract arguments of the target event.

To be more specific, at the $i$-th layer of encoder or decoder, we equip the self-attention block with a set of learnable prefix tokens oriented towards event instances, \(\text{Prefix}_{i}^{\textit{ins}} \in \mathbb{R}^{len_{\textit{ins}} \times m}\), and event templates, \(\text{Prefix}_{i}^{\textit{tem}} \in \mathbb{R}^{len_{\textit{tem}} \times m}\). Here, \(len_{\textit{ins}}\) is the length of \(\text{Prefix}_{i}^{\textit{ins}}\), \(len_{\textit{tem}}\) is the length of \(\text{Prefix}_{i}^{\textit{tem}}\), and \(m\) corresponds to the hidden dimension. We also attempt to equip the cross-attention block in each layer of the decoder with these prefix tokens, but observe almost no improvement in performance. 
These additional key and value matrices will be concatenated with the original key and value matrices in the attention block. When calculating dot-product attention, the query at each position will be influenced by these prefix tokens.

\subsubsection{Event-Guided Adaptive Gating mechanism}
\label{Event-Guided Adaptive Gating mechanism}
% Not all information within the prefix benefits argument extraction of the target event.
Some relevant events may provide useful cues to the target event and irrelevant event information might hinder model performance.
To connect other relevant events to the target event and thus adaptively capture relevant information from the prefix, inspired by \citet{liu-etal-2023-enhancing-document}, we utilize the target event's trigger and type to score each prefix token in $\text{Prefix}_{i}^{ins}$ and $\text{Prefix}_{i}^{tem}$ through gated weight assignment, respectively. For example, in the $i$-th layer, the weight distribution of tokens in $\text{Prefix}_{i}^{ins}$ is calculated as:
\begin{equation}
a_{i}^{ins} = \text{sigmoid}(h_{i-1}^{trig} W^{ins}_{i}),
\label{eq1}
\end{equation}
where $h_{i-1}^{trig}$ is the hidden state of the event trigger from the previous layer obtained through average pooling, and $W^{ins}_{i} \in \mathbb{R}^{m \times len_{ins}}$ corresponds to the parameters to be learned. Here, the prefix weight $a_{i}^{ins}$ captures the relevance of other event instances to the given instance, and based on this, we can further capture the relevant information within prefix.

Additionally, we also follow \citet{zhang-etal-2023-towards-adaptive} to introduce a learnable scaling factor $\lambda_{i}^{ins}$ to balance the information provided by the original input and the prefix. We obtain the guided instance-oriented prefix  representation $\text{Prefix}_{i}^{ins'}$:
\begin{equation}
\text{Prefix}_{i}^{ins'} = \lambda_{i}^{ins}a_{i}^{ins} \odot \text{Prefix}_{i}^{ins},
\end{equation}

\noindent
where $\odot$ is the element-wise multiplication. Similarly, we  derive the guided template-oriented prefix  representation $\text{Prefix}_{i}^{tem'}$:
\begin{equation}
a_{i}^{tem} = \text{sigmoid}(h_{i-1}^{type} W^{tem}_{i}),
\label{a_tem}
\end{equation}
\begin{equation}
\text{Prefix}_{i}^{tem'} = \lambda_{i}^{tem}a_{i}^{tem} \odot \text{Prefix}_{i}^{tem}, 
\label{eq3}
\end{equation}

\noindent
where $h_{i-1}^{type}$ is the hidden state obtained through average pooling of the event type from the previous layer, and $W^{tem}_{i} \in \mathbb{R}^{m \times len_{tem}}$ is learnable parameters. The prefix weight $a_{i}^{tem}$ captures the relevance of other event templates to the given template.

Finally, these event-guided prefixes can more effectively provide relevant event instances and templates as cues for the target event. The new self-attention module with event information intervention is formalized as:
\begin{equation*}
H \leftarrow \text{LayerNorm}(H^{'} + H),
\end{equation*}
\begin{equation}
H^{'} = \text{MHSA}(\text{Prefix}_i^{'
} \oplus H)_{|\text{Prefix}_i^{'}|:|\text{Prefix}_i ^{'} \oplus H|},
\end{equation}
where $\text{MHSA}(\cdot)$ denotes the multi-head self-attention mechanism, $\text{Prefix}i^{'}$ represents either $\text{Prefix}{i}^{ins'}$ or $\text{Prefix}{i}^{tem'}$, $H$ refers to the hidden state, and ${b:c}$ denotes the slicing operation on the $seq\_len$ dimension from $b$ to $c$.

% "this refers to the slicing operation performed on the sequence length dimension. let's illustrate with an example: the shape of tensor A is [bsz, prefix_len + sequence_len, hidden_size], after slicing on the sequence length dimension from |prefix_len| to |prefix_len + sequence_len|, the shape of tensor A' becomes [bsz, sequence_len, hidden_size].  "

\subsubsection{Encoding with Prefixes}
\label{Encoding with Prefixes}
To ensure a fair comparison and to visually demonstrate the effects of our prefixes, we adopt the encoding process from prior work \cite{he2023revisiting} for encoding context instance and event template, and augment it with our prefixes. With both relevant event instances and templates as cues 
% and learning through analogy
, model can better understand the target event from context.

Given input context $X = \{x_1, x_2, \ldots, x_n\}$ and the event trigger $x_{trig}$ 
% triggering event type $e$
, we use text markers \textbf{<t>} and \textbf{</t>} as special tokens and then insert them before and after $x_{trig}$ in $X$, respectively.
\begin{equation*}
    \widetilde{X} = \{x_1, x_2, \ldots, \textbf{<t>} x_{trig} \textbf{</t>}, \ldots, x_n\}.
\end{equation*}
Then, we input the processed $\widetilde{X}$ into the encoder and the decoder to obtain the context representation $H_{\widetilde{X}}$:
\begin{equation}
H_{\widetilde{X}}^{enc} = \text{Encoder}(\text{Prefix}_{enc}^{ins}, \widetilde{X}),
\label{eq6}
\end{equation}
\begin{equation}
H_{\widetilde{X}} = \text{Decoder}(\text{Prefix}_{dec}^{ins}, H_{\widetilde{X}}^{enc}), 
\label{hx decoder}
\end{equation}
where $\text{Prefix}_{enc}^{ins}$ and $\text{Prefix}_{dec}^{ins}$ are the instance-oriented prefix representations $\text{Prefix}_{i}^{ins}$ from the encoder and decoder, respectively. 

For the target event template $\mathcal{T}_{e}$, we input it into the encoder and the decoder to obtain the template representation $H_{\mathcal{T}_{e}}$:
\begin{equation}
H_{\mathcal{T}_{e}}^{enc} = \text{Encoder}(\text{Prefix}_{enc}^{tem}, \mathcal{T}_{e}),
\end{equation}
\begin{equation}
H_{\mathcal{T}_{e}} = \text{Decoder}(\text{Prefix}_{dec}^{tem}, H_{\mathcal{T}_{e}}^{enc}, H_{\widetilde{X}}^{enc}), 
\label{eq7}
\end{equation}
where $\text{Prefix}_{enc}^{tem}$ and $\text{Prefix}_{dec}^{tem}$ are the template-oriented prefix representations $\text{Prefix}_{i}^{tem}$ from the encoder and decoder, respectively. 

\subsubsection{Span Selection}
\label{Span Selection}
With the template representation $H_{\mathcal{T}_{e}}$ obtained through Eq.~\ref{eq7}, for the $k$-th role slot in the template, we can obtain its representation $h_{s_k}$ and span selector \(\phi_{k} = \{\phi_{k}^{start}, \phi_{k}^{end}\}\) \cite{ma2022prompt, he2023revisiting}:
\begin{equation*}
    \phi_{k}^{start} = h_{s_{k}} \odot w^{start},
\end{equation*}
\begin{equation}
    \phi_{k}^{end} = h_{s_{k}} \odot w^{end},
\end{equation}
where $w^{start}$ and $w^{end}$ are learnable parameters. Subsequently, the span selector $\phi_{k}$ queries the context to extract at most one argument span $(s_{k}, e_{k})$ for role slot $k$:  
\begin{equation*}
\text{logit}_{k}^{start} = \phi_{k}^{start} H_{\widetilde{X}} \in \mathbb{R}^{n},
\end{equation*}
\begin{equation*}
\text{logit}_{k}^{end} = \phi_{k}^{end} H_{\widetilde{X}} \in \mathbb{R}^{n},
\end{equation*}
\begin{equation*}
\text{score}_{k}(l, r) = \text{logit}_{k}^{start}(l) + \text{logit}_{k}^{end}(r),
\end{equation*}
\begin{equation}
(s_{k}, e_{k}) = \argmax_{(l, r) \in \mathcal{C}} \text{score}_{k}(l, r)
\end{equation}
where candidate span set $\mathcal{C} = \{(l,r) \mid (l, r)  \in n^2, 0 < r - l < MSL\} \cup \{(0,0)\} $. $MSL$ denotes the max span length.
The loss function is defined as:
\begin{equation*}
P_{k}^{start} = \text{Softmax}(\text{logit}_{k}^{start}),
\end{equation*}
\begin{equation*}
P_{k}^{end} = \text{Softmax}(\text{logit}_{k}^{end}),
\end{equation*}
\begin{equation}
\mathcal{L} = - \sum_{X} \sum_{k} [\log P_{k}^{start}(s_{k}) + \log P_{k}^{end}(e_{k})],
\end{equation}
where \(X\) ranges over all contexts in dataset and \(k\) ranges over all role slots in the template for \(X\), and \((s_{k}, e_{k})\) is the golden span for role slot \(k\). For a role slot that does not relate to any argument, it is assigned the empty span \( (0,0) \).
\section{Experiments}

\subsection{Experimental Setup}

\begin{table*}[t]
\centering
\resizebox{0.95\textwidth}{!}{
\begin{tabular}{l|l|l|cccccccc}
\toprule
\multirow{2}{*}{Category} & \multirow{2}{*}{Model} & \multirow{2}{*}{PLM} & \multicolumn{2}{c}{ACE05} & \multicolumn{2}{c}{RAMS} & \multicolumn{2}{c}{WIKIEVENTS}& \multicolumn{2}{c}{MLEE} \\
\cline{4-11}
& & & Arg-I & Arg-C & Arg-I & Arg-C & Arg-I & Arg-C & Arg-I & Arg-C \\ 
\midrule
\multirow{2}{*}{Retrieval-based} &R-GQA  \cite{du-ji-2022-retrieval} & BART & 75.6 & 72.8 & - & - & - & - & - & - \\
&AHR \cite{ren-etal-2023-retrieve} & T5 & - & - & 54.6 & 48.4 & 69.6 & 63.4 & - & - \\  \midrule
\multirow{4}{*}{Templated-based}&BART-Gen \cite{li-etal-2021-document} & RoBERTa & 69.9 & 66.7 & 51.2 & 47.1 & 66.8 & 62.4 & 71.0 & 69.8 \\
&DEGREE \cite{hsu-etal-2022-degree} & BART & 75.6 & 73.0 & - & - & - & - & - & - \\
&PAIE \cite{ma2022prompt} & BART & 75.7 & 72.7 & 56.8 & 52.2 & 70.5 & 65.3 & 72.1 & 70.8 \\
&TabEAE \cite{he2023revisiting} & RoBERTa & 75.5 & 72.6 & 57.0 & 52.5 & 70.8 & 65.4 & 71.9 & 71.0 \\ \midrule
\multirow{3}{*}{Prefix-based}
&APE \cite{zhang2023overlap} & BART & 75.3 & 72.9 & 56.3 & 51.7 & 70.6 & 65.8 & - & - \\
&AMPERE \cite{hsu2023ampere} & BART & 76.0 & \underline{73.4} & - & - & - & - & - & - \\
&SPEAE \cite{nguyen2023contextualized} & BART & - & - & \underline{58.0} & \underline{53.3} & \underline{71.9} & \underline{66.1} & - & - \\ 
\midrule
Retrieval-based &DRAAE (Ours)  & RoBERTa & \underline{76.1} & 73.2 & 57.5  &  53.0 & 71.0 & 66.0 & \underline{72.8}  & \underline{71.7} \\
Prefix-based &DEGAP (Ours) & RoBERTa & \textbf{76.6} & \textbf{74.4} & \textbf{58.5} & \textbf{54.2} & \textbf{72.2} & \textbf{67.1} & \textbf{74.0} & \textbf{73.4} \\
\bottomrule
\end{tabular}
}
\caption{The overall performance of all methods. Models in the PLM column are of large-scale (with 24 Transformer layers). The highest scores are highlighted in bold, and the second-highest scores are underlined.}
\label{tab1}
\end{table*}

\textbf{Datasets}. We conduct experiments on four benchmark datasets, specifically including ACE05 \cite{doddington2004automatic}, RAMS \cite{ebner-etal-2020-multi}, WIKIEVENTS \cite{li-etal-2021-document}, and MLEE \cite{pyysalo2012event}. ACE05 is a sentence-level dataset, while the other three are document-level datasets. We provide detailed introductions to these datasets in Appendix~\ref{datasets}.

\vspace{3pt}
\noindent
\textbf{Evaluation Metrics.} We adopt two evaluation metrics: (1) Argument Identification F1 score (\textbf{Arg-I}): A predicted argument for an event is correctly identified if its boundary matches any golden arguments of the event. (2) Argument Classification F1 score (\textbf{Arg-C}): a predicted argument of an event is correctly classified only if its boundary and role type both match the ground truth. The final reported performance is averaged over five random runs.

\vspace{3pt}
\noindent
\textbf{Implementation details}. 
% We implement DEGAP using PyTorch and conduct our experiments on a single NVIDIA GeForce RTX 3090 24G. For simplicity, we randomly initialize the embeddings of the prefix. Following TabEAE's setting \cite{he2023revisiting}, we use the first 17 layers of RoBERTa-large \cite{liu2019roberta} to instantiate the encoder, and the weights of the remaining 7 layers to initialize the weight of self-attention layers and feedforward layers of the decoder. We also randomly initialize the weight of cross-attention parts of the decoder. The optimization of our model employs the AdamW optimizer \cite{loshchilov2017decoupled} equipped with a linear learning rate scheduler. 
Please refer to Appendix~\ref{hyperparameters settings} for details of implementation.

\vspace{3pt}
\noindent
\textbf{Baselines}. 
% We compare our DEGAP with several state-of-the-art models across three categories: (1) \textbf{Classification-based model}: TSAR \cite{xu-etal-2022-two}, SCPRG \cite{liu2023enhancing}; (2) \textbf{Generation-based model}: BART-Gen \cite{li-etal-2021-document}; DEGREE \cite{hsu-etal-2022-degree}, AMPERE \cite{hsu2023ampere},  APE \cite{zhang2023overlap}, AHR \cite{ren-etal-2023-retrieve}. (3) \textbf{QA-based model}: EEQA \cite{du-cardie-2020-event}, PAIE \cite{ma2022prompt}, SPEAE \cite{nguyen2023contextualized}, TabEAE \cite{he2023revisiting}, R-GQA \cite{du-ji-2022-retrieval}. 
Details about the baselines are listed in Appendix~\ref{baselines}. For a comprehensive comparison, we also conduct a comparative analysis between DEGAP and LLMs in Appendix~\ref{Performance Comparison with LLMs}, showing that DEGAP can outperform the latest LLMs for EAE.

We also compare our method with its variant, the \textbf{D}ual \textbf{R}etrieval-\textbf{A}ugmented \textbf{A}rgument \textbf{E}xtraction model (DRAAE), which only retrieves relevant event instances and templates as cues to guide the model, instead of using the prefixes we designed. We provide further implementation details of this model in Appendix~\ref{The implementation details of DRAAE}.

\subsection{Overall Performance}

Table~\ref{tab1} presents the performance of all baselines and our method across all datasets. Our method %significantly 
outperforms all the baselines and achieves new SOTA results on the four datasets. Specifically, our method improves the Arg-C F1 score by 1.0\%, 0.9\%, 0.9\%, and 1.9\% on ACE05, RAMS, WIKIEVENTS, and MLEE, respectively. This demonstrates the efficacy of our method in handling both sentence-level EAE and document-level EAE tasks. Moreover, we can conclude:

(1) \textit{Augmenting input with both relevant event instances and templates is beneficial for extracting arguments}. Our DRAAE and DEGAP outperform previous retrieval-based and template-based methods, which only utilize event instances or templates to provide insufficient guidance for model. Our approach can fully combine these beneficial cues to better understand the target event from context.

(2) \textit{Our event-guided prefix can effectively coping with the specific informational needs
of different events, which is superior to previous prefixes.} Compared to previous prefix-based methods, such as AMPERE and SPEAE, which utilize auxiliary semantic information to augment their task-general prefixes, our method does not rely on these modules and still surpasses them to varying degrees.
This reveals that by using EGAG to connect other relevant events to the target event, our method can effectively utilize relevant information in the prefix and discard irrelevant information.
% This reveals that our event-guided prefixes are sufficient to cope with the specific informational needs of different events, leveraging their advantages.

(3) \textit{Our prefixes can utilize relevant event instances and templates more sufficiently and efficiently to boost performance compared to retrieval.} 
As shown in Table~\ref{tab1}, 
Our DEGAP outperforms DRAAE on all datasets. This is because  1) DRAAE uses a pre-trained retriever with frozen parameters to retrieve event instances related to the target event, but fails to find the most relevant events, %resulting in insufficient retrieval, 
and 2) DRAAE retrieves and encodes independently, leading to a poor local optimum for each of the steps. In contrast, our prefixes are trained and updated along with model parameters and can fully learn the semantic information of different event instances and templates, avoiding these issues.  Moreover, 
the efficiency analysis in Appendix~\ref{Efficiency Analysis} indicates that %due to our DEGAP not requiring retrieval, it 
DEGAP has a faster inference speed due to the 
removal of retrieval.

\section{Analysis}

In this section, we conduct a comprehensive study to analyze the reasons for the performance improvement of our method, and to validate the significance of our proposed design and the effectiveness of the main components.

\subsection{Why is DEGAP Effective?}

We explore from two perspectives: 1) domain transfer and 2) handling complex contextual semantics, to investigate whether the performance gains in our method are due to DEGAP providing the model with relevant event instances and event templates as cues. We consider the base model TabEAE, which primarily differs from our model in whether dual prefixes are used.

\vspace{3pt}
\noindent
\textbf{Domain Transfer}. We assume that DEGAP can use relevant event information from the source domain as cues and learn by analogy to better understand new types of events in the target domain. To this end, we examine the model performance when facing domain transfer. Specifically, we select four different domain datasets: ACE05, RAMS, WIKIEVENTS, and MLEE, train the model on one domain dataset \texttt{(src)}, and then test it on all domain datasets \texttt{(tgt)}. 
The results are shown in Table~\ref{domain_tansfer} in Appendix~\ref{appendix Domain Transfer}. Compared to the results w/o domain transfer, both our method DEGAP and the baseline TabEAE exhibit a sharp decline under different \texttt{src-tgt} transfer settings, where the baseline's performance even falls below 10\% in some cases (e.g., WIKI $\Rightarrow$ MLEE). This indicates that extracting arguments from new types of events is highly challenging for the model. However, encouragingly, our method DEGAP consistently outperforms the baseline in different \texttt{src-tgt} transfer settings. This fully confirms that our hypothesis is correct.

\vspace{3pt}
\noindent
\textbf{Handing Complex Contextual Semantics}. We believe that by fully utilizing relevant event instances and event templates to guide the model, our method excels in identifying the target event from complex contextual semantics. To this end, we explore the ability of models to handle complex semantics from two perspectives: the number of events and the degree of event relevance. \textbf{(1) Number of events.} Multiple events in the same context often exhibit complex semantic connections. We measured the performance of the models on instances whose context contains different numbers of events. As shown in Table~\ref{multi event} in Appendix~\ref{appendix Handing Complex Contextual Semantics}, compared to the baseline TabEAE, our method shows consistent %significant 
improvements in performance across all datasets, especially in instances that contain multiple events within the same context. \textbf{ (2) Degree of event relevance.} If two events in the context share arguments, their semantic connection is often more complex. We measured the performance of the model in extracting arguments from overlapping events (events with shared arguments). Table~\ref{overlap event} shows that our method achieves the best performance across all datasets, particularly gaining greater improvements in handling overlapping events. Experimental results support our conclusion

\subsection{Why are Dual Prefixes Necessary?}
\label{Why are Dual Prefixes Necessary?}
\begin{table}[t]
    \centering
    \resizebox{1.0\columnwidth}{!}{
    \begin{tabular}{c|l|cccc}
    \toprule
     No. &  Model  &  ACE05 & RAMS & WIKI & MLEE \\
    \midrule
     1 &  DEGAP &  74.4 & 54.2 & 67.1 & 73.4 \\
      \midrule
     2 &  DEGAP-CCT & 73.6 & 53.7 & 66.6 & 72.3 \\
    3 &   DEGAP-SP & 73.9 & 53.4 & 66.8 & 72.6 \\
      \midrule
    4 &   only IOP (w/o TOP) & 73.8 & 52.9	& 66.4	& 72.6  \\
     5 &  only TOP ( w/o IOP) & 73.6 & 52.7 & 66.3 & 72.7  \\
    6 &   w/o DEGAP & 72.8 & 52.0	& 65.7	& 71.2 \\
      \midrule
    7 &  DEGAP-TST & 73.9	& 53.0	& 66.6	& 72.8 \\
    \bottomrule
    \end{tabular}
    }
    \caption{Ablations (Arg-C F1 score) on dual prefixes. Note that all models here are equipped with our proposed event-guided adaptive gating mechanism.}
    \label{tab2}
\end{table}

A core question is why we need to design separate prefixes for event instances and event templates. To explore whether the design of dual prefixes is necessary in our method, we compare DEGAP with two single-prefix variants:
% \begin{itemize}
%     \item \textbf{DEGAP-CCT} (\textbf{C}oncatenate \textbf{C}ontext and \textbf{T}emplate): Instead of encoding the context and template separately, this variant directly concatenates them and then inputs the combined sequence into the encoder, which is equipped with a shared prefix. For a fair comparison, we use RoBERTa-large as the encoder.
%     \item \textbf{DEGAP-SP} (\textbf{S}hared \textbf{P}refix): This variant follows the same encoding process as DEGAP but replaces instance-oriented and template-oriented prefixes with a shared prefix.
% \end{itemize}

(1) \textbf{DEGAP-CCT} (\textbf{C}oncatenate \textbf{C}ontext and \textbf{T}emplate): Instead of encoding the context and template separately, this variant directly concatenates them and then inputs the combined sequence into the encoder, which is equipped with a shared prefix. For a fair comparison, we use RoBERTa-large as the encoder.

(2) \textbf{DEGAP-SP} (\textbf{S}hared \textbf{P}refix): This variant follows the same encoding process as DEGAP but replaces instance-oriented and template-oriented prefixes with a shared prefix.

Moving forward, we further investigate the effects of instance-oriented prefix (IOP) and template-oriented prefix (TOP). Finally, we consider DEGAP-TST (event-\textbf{T}emplate-\textbf{S}pecific \textbf{T}OP), where each event template has its specific TOP. Therefore, each TOP solely learns and provides semantic information specific to the target event template, leading to insufficient interaction between event templates.  Table~\ref{tab2} shows the results.

From the table, we observe that compared to our DEGAP (line 1), the single-prefix variants DEGAP-CCT and DEGAP-SP (lines 2-3) show performance declines to varying degrees across all datasets. 
This may be because the prefix simultaneously learns the semantic information of event instances and templates, leading to confusion between this information and failing to accurately provide sufficient guidance for model when encoding context and event template. In contrast, dual prefixes separately learn semantic information of event instance and template, which can better provide %correspondingly sufficient beneficial 
useful cues for model when encoding context and event template.

Additionally, compared to the baseline w/o DEGAP (line 6), introducing IOP or TOP separately will result in a performance increase on all datasets (lines 4-5), and the performance gains by introducing them at the same time exceed the performance improvement  by introducing any one module alone (line 1). This indicates that it is necessary to utilize IOP and TOP to respectively provide the model with relevant event instance information and event template information as cues for inference.  Moreover, IOP and TOP can offer complementary guidance to the model from the event mention-level and ontology level, respectively.

Finally, an interesting phenomenon is that despite DEGAP-TST introducing more parameters than DEGAP, its performance declines across all datasets (line 7). This confirms the importance of considering the potential connections between templates, where some relevant templates can usually provide useful information to the target event.

\subsection{Is Event-Guided Adaptive Gating mechanism Necessary?}

\begin{table}[t]
    \centering
    \resizebox{0.5\textwidth}{!}{
    \begin{tabular}{c|l|cccc}
    \toprule
    No. &  Model  &  ACE05 & RAMS & WIKI & MLEE \\
    \midrule
    1 &   DEGAP &  74.4 & 54.2 & 67.1 & 73.4 \\
      \midrule
   2 &    w/o EGAG & 73.4	& 53.3	& 66.4	& 72.3 \\
    3 &   DEGAP-S & 73.9	& 53.6 & 66.8 & 72.6 \\ 
    % \midrule
    % 4 & w/o DEGAP & 72.8 & 52.0	& 65.7	& 71.2 \\
    \bottomrule
    \end{tabular}
    }
    \caption{The study (Arg-C F1 score) of event-guided adaptive gating mechanism.}
    \label{tab3}
\end{table}

To explore the importance of the Event-Guided Adaptive Gating mechanism (EGAG), we compare DEGAP with the following variants:
% \begin{itemize}
%     \item \textbf{w/o EGAG}: To investigate whether the performance improvement of DEGAP is due to the additional parameters introduced (gating weights and scaling factors) or because event-guided prefixes can better provide model with relevant event information as cues, we designed the variant DEGAP w/o EGAG. For a fair comparison, we expanded the prefix length to 1.5 times to align its parameter count with DEGAP.
%     \item \textbf{DEGAP-S}: To validate the importance of using the target event to guide prefix, we follow \citet{zhang-etal-2023-towards-adaptive} to directly use the sentence representation in the previous layer as  $h_{i-1}^{trig}$ in Eq.~\ref{eq1} and $h_{i-1}^{type}$ in Eq.~\ref{a_tem}, respectively.
% \end{itemize}

(1) \textbf{w/o EGAG}: To investigate whether the performance improvement of DEGAP is due to the additional parameters introduced (gating weights and scaling factors) or because event-guided prefixes can better provide model with relevant event information as cues, we designed the variant DEGAP w/o EGAG. For a fair comparison, we expanded the prefix length to 1.5 times to align its parameter count with DEGAP.

(2) \textbf{DEGAP-S}: To validate the importance of using the target event to guide prefix, we follow \citet{zhang-etal-2023-towards-adaptive} to directly use the sentence representation in the previous layer as  $h_{i-1}^{trig}$ in Eq.~\ref{eq1} and $h_{i-1}^{type}$ in Eq.~\ref{a_tem}, respectively.

Table~\ref{tab3} shows that removing EGAG results in a noticeable decline in Arg-C F1 scores across the four datasets. This phenomenon indicates that utilizing EGAG to connect other relevant events to the target one is very necessary, as it allows our method to further effectively utilize relevant information in the prefix and discard irrelevant information. Additionally, DEGAP-S also shows a decrease in Arg-C F1 scores across the four datasets. This may be because it uses only the first input token “\texttt{<s>}” to calculate $h_{i-1}^{trig}$ and $h_{i-1}^{type}$, which is insufficient to connect other relevant events to the target event and thus capture relevant information from the prefix. However, compared to the model w/o EGAG, DEGAP-S shows improved performance across all four datasets. This further confirms the importance of utilizing EGAG to connect other relevant events to the target event.

% \begin{figure}[t]
% \centering
% \subfigure{
% \begin{minipage}[t]{0.50\linewidth}
% \centering
% \includegraphics[width=\linewidth]{con_att.png}
% \end{minipage}%
% }%
% \subfigure{
% \begin{minipage}[t]{0.47\linewidth}
% \centering
% \includegraphics[width=\linewidth]{pro_att.png}
% \end{minipage}%
% }%
%  \caption{The weight distribution of prefix tokens, which is guided by the target event using the event-guided adaptive gating mechanism.}
%     \label{prefix token weight}
% \end{figure}

% \begin{figure}[t]
% \centering
% \subfigure{
% \begin{minipage}[t]{0.49\linewidth}
% \centering
% \includegraphics[width=\linewidth]{len_con.png}
% \end{minipage}%
% }%
% \subfigure{
% \begin{minipage}[t]{0.49\linewidth}
% \centering
% \includegraphics[width=\linewidth]{len_pro.png}
% \end{minipage}%
% }%
% \caption{Results under different prefix lengths.}
%     \label{prefix len}
% \end{figure}

\textbf{Attention Visualization.} We further demonstrate the effects of EGAG visually on the ACE05 dataset. We randomly select a sample and visualize the weight distribution of prefix tokens in  IOP and TOP. As shown in Fig.~\ref{prefix token weight} in Appendix~\ref{appendix attention visualization}, the target event has different levels of attention to the prefixes of different layers. More specifically, even within the same layer, the attention to different prefix tokens varies. This observation confirms that EGAG can adaptively capture the relevance of other event to the given one and further extract relevant information from the prefix tokens at each layer.

\textbf{Prefix length.} Moreover, We analyze the impact of prefix length on model performance in Appendix~\ref{Prefix length}.
% "Note that instance/template-oriented prefix has learned semantic information from all event instances/templates. Moreover, not all event instances and templates are relevant to the target event. We aim to leverage only those instances and templates that are beneficial to the target event. Thus, the visual attention mechanism can be seen as selecting relevant and useful event instances/templates. Consequently, the regular attention visualization results appear to be somewhat unreasonable."

% \textbf{Prefix length}. We further analyze the impact of prefix length on model performance. Intuitively, a longer prefix implies more trainable parameters, thus enabling it to provide more expressive guidance for the model regarding relevant semantic information. As shown in Fig.~\ref{prefix len}, as the prefix length increases, the model performance tends to improve, followed by a slight decline. We speculate that this is because too long prefixes provide redundant semantic information, making it difficult for the model to focus on extracting the target event.

\section{Conclusion}

In this work, we introduce DEGAP. Based on observations of existing research and comprehensive experimental analysis, we show that prefix can effectively learn semantic information from different event instances and templates, and allow them to interact fully. Additionally, we introduce an event-guided adaptive gating mechanism to connect other relevant events to the target event and thus capture relevant information from the prefix.
% that guides prefixes to adapt to adaptively
% different events to fully leverage their advantages. 
Finally, these event-guided prefixes provide relevant event instances and template information as cues for the target event.
Extensive experiments demonstrate the validity of our newly introduced methods. In the future, we plan to integrate other useful auxiliary information into prefixes for solving EAE.

\section*{Limitations}

We summarize the limitations of our method as follows:

(1) Although experimental results indicate that our method exhibits profound performance, our designed prefixes lack sufficient interpretability compared to methods that directly retrieve relevant event instances and templates to enhance performance (e.g., DEGAP vs DRAAE). Additionally, compared to other methods, our approach
requires additional time costs to find suitable prefix
lengths for both IOP and TOP.

(2) Due to limited computational resources, we only equipped our prefixes for use on PLMs. In fact, some other works have successfully integrated these efficient prefix parameters into LLMs (e.g., Llama 2 \cite{touvron2023llama}) and demonstrated excellent results \cite{zhang2023llama}. How to effectively combine our prefixes with these LLMs for solving the EAE task remains an open research question.

We will attempt to address these issues in future work.

\bibliography{anthology,custom}
\bibliographystyle{acl_natbib}

\clearpage

\appendix

\section{Why is DEGAP Effective?}
\label{appendix why is degap effective}

\begin{table*}[t!]
    \centering
    \resizebox{1.0 \textwidth}{!}{
    \begin{tabular}{l|cccc|cccc|cccc|cccc|c}
    \toprule
        % \multirow{3}{*}{Model} & \multicolumn{3}{c|}{ACE05} & \multicolumn{3}{c|}{RAMS} & \multicolumn{3}{c|}{WIKIEVENTS} & \multicolumn{3}{c|}{MLEE} & \multirow{3}{*}{avg} \\   \cline{2-13}
         \multirow{3}{*}{Model}  & ACE05 & ACE05 & ACE05 & ACE05 & RAMS & RAMS & RAMS & RAMS & WIKI & WIKI & WIKI & WIKI & MLEE & MLEE & MLEE & MLEE & \multirow{3}{*}{Avg}\\   
         & $\Downarrow$ & $\Downarrow$ & $\Downarrow$ & $\Downarrow$ & $\Downarrow$ & $\Downarrow$ & $\Downarrow$ & $\Downarrow$ & $\Downarrow$ & $\Downarrow$ & $\Downarrow$ & $\Downarrow$ &
         $\Downarrow$ & $\Downarrow$ & $\Downarrow$ & $\Downarrow$ \\
           & ACE05 & RAMS & WIKI & MLEE & RAMS & ACE05 & WIKI & MLEE & WIKI & ACE05 & RAMS & MLEE & MLEE & ACE05 & RAMS & WIKI \\
    \midrule
    TabEAE & 72.6 & 20.1 & 41.2 & 5.4 & 52.5 & 32.3 & 20.1 & 19.6 & 65.4 & 48.7 & 32.6 & 9.6 & 71.0 & 8.5 & 14.4 & 9.9 & 32.7 \\
    DEGAP & \textbf{74.4} & \textbf{23.8} & \textbf{44.8} & \textbf{11.5} & \textbf{54.2} & \textbf{33.9} & \textbf{27.6} & \textbf{25.7} & \textbf{67.1} & \textbf{50.1} & \textbf{34.4} & \textbf{29.3} & \textbf{73.4} & \textbf{21.9} & \textbf{19.1} & \textbf{22.5} & \textbf{38.4} \\
    \bottomrule
    \end{tabular}
    }
    \caption{Performance (Arg-C F1 score) under different \texttt{src}$\Rightarrow$\texttt{tgt} settings. We train the model on the \texttt{src} dataset and test it on the \texttt{tgt} dataset. The numbers in the Avg column represent the average of the numbers under all \texttt{src}$\Rightarrow$\texttt{tgt} settings. WIKIEVENTS is abbreviated as WIKI (the same below).}
    \label{domain_tansfer}
\end{table*}

\begin{table*}[t!]
    \centering
    \resizebox{1.0\textwidth}{!}{
    \begin{tabular}{l|cccccccc}
    \toprule
    \multirow{2}{*}{Model} & \multicolumn{2}{c}{ACE05} & \multicolumn{2}{c}{RAMS} & \multicolumn{2}{c}{WIKI} & \multicolumn{2}{c}{MLEE} \\  \cline{2-9}
        &  S$_{[\textbf{185}]}$ &  M$_{[\textbf{218}]}$ & S$_{\textbf{[587]}}$ & M$_{\textbf{[284]}}$ & S$_{\textbf{[114]}}$ & M$_{\textbf{[251]}}$ & S$_{\textbf{[175]}}$ & M$_{\textbf{[2025]}}$ \\
    \midrule
    TabEAE &  71.2 &	73.8	& 52.8	& 51.6	& 65.3	& 65.5 &	79.3	& 70.3  \\
    DEGAP & \textbf{72.7} (1.5) & \textbf{75.8} (2.0)& \textbf{54.4} (1.6)& \textbf{54.0} (2.4)& \textbf{66.5} (1.2)& \textbf{67.3} (1.8)& \textbf{81.1} (1.8)& \textbf{72.8} (2.5)\\
    \bottomrule
    \end{tabular}
    }
    \caption{Performance (Arg-C F1 score) on instances with different numbers of events in context. The numbers in square brackets indicate the number of supporting events in the test set. \textbf{S}: Single event. \textbf{M}: Multiple events.}
    \label{multi event}
\end{table*}

\begin{table*}[t!]
    \centering
    \resizebox{1.0\textwidth}{!}{
    \begin{tabular}{l|cccccccc}
    \toprule
    \multirow{2}{*}{Model} & \multicolumn{2}{c}{ACE05} & \multicolumn{2}{c}{RAMS} & \multicolumn{2}{c}{WIKI} & \multicolumn{2}{c}{MLEE} \\  \cline{2-9}
        & N-O$_{\textbf{[319]}}$ & Overlap $_{\textbf{[84]}}$ & N-O$_{\textbf{[690]}}$ & Overlap$_{\textbf{[181]}}$ & N-O$_{\textbf{[296]}}$ & Overlap$_{\textbf{[69]}}$ & N-O$_{\textbf{[1466]}}$ & Overlap$_{\textbf{[734]}}$ \\
    \midrule
    TabEAE & 71.1 & 78.6 & 51.6 & 55.6 & 65.7 & 64.4 & 75.4 & 65.8   \\
    DEGAP & \textbf{72.1} (1.0) & \textbf{80.5} (1.9)& \textbf{52.9} (1.3)& \textbf{59.5} (3.9)& \textbf{67.3} (1.6)& \textbf{66.7} (2.3)& \textbf{76.9} (1.5)& \textbf{68.5} (2.7)\\
    \bottomrule
    \end{tabular}
    }
    \caption{Results (Arg-C F1 score) of extracting overlapping (events with shared arguments) and non-overlapping events. Numbers in square brackets indicate the number of supporting events in the test set. \textbf{N-O}: Non-overlapping.}
    \label{overlap event}
\end{table*}

\subsection{Domain Transfer.} 
\label{appendix Domain Transfer}

We examine the model performance when facing domain transfer. Table~\ref{domain_tansfer} shows the Arg-C F1 score under different \texttt{src}$\Rightarrow$\texttt{tgt} settings. We can find that our method DEGAP consistently outperforms the baseline TabEAE in different \texttt{src-tgt} transfer settings. This may be because DEGAP can use relevant event information from the source domain as cues and learn by analogy to better understand new types of events in the target domain. 
As TabEAE doesn't have this capability, it shows a huge performance gap compared to our DEGAP in some settings (e.g., WIKI $\Rightarrow$ MLEE).

\subsection{Handing Complex Contextual Semantics.} 
\label{appendix Handing Complex Contextual Semantics}

We explore the ability of models to handle complex semantics from two perspectives: the number of events and the degree of event relevance. Table~\ref{multi event} presents the Arg-C F1 score on instances with different numbers of events in context, and Table~\ref{overlap event} shows the results of extracting overlapping (events with shared arguments) and non-overlapping events. 
We can infer that the performance improvement of DEGAP is more attributed to event instances that contain multiple events within the same context and overlapping events. This may be because by fully utilizing relevant event instances and event templates to guide the model, our DEGAP can better identify the target event from complex contextual semantics.

\section{Attention Visualization}
\label{appendix attention visualization}

\begin{figure}[t]
\centering
\subfigure{
\begin{minipage}[t]{0.50\linewidth}
\centering
\includegraphics[width=\linewidth]{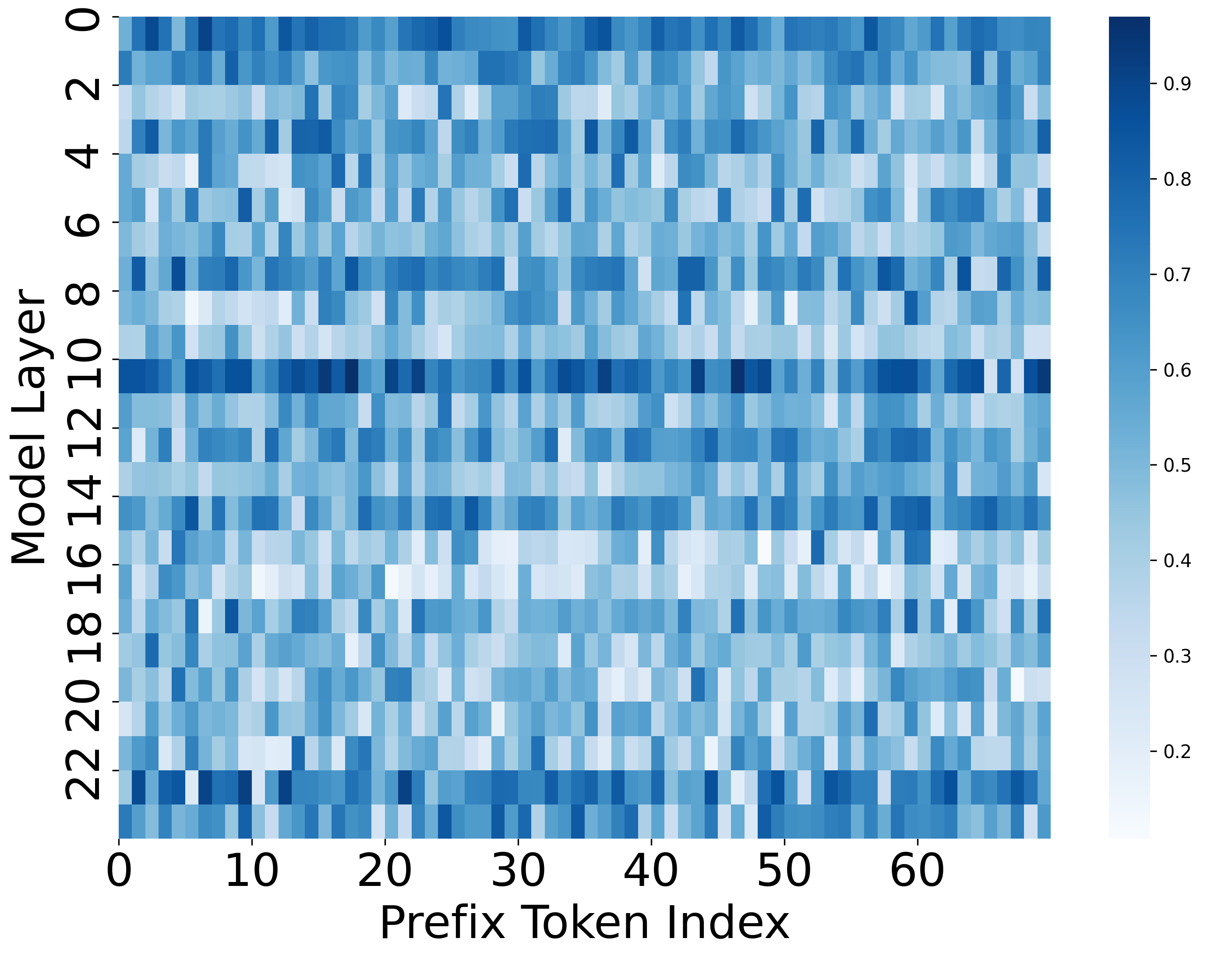}
\end{minipage}%
}%
\subfigure{
\begin{minipage}[t]{0.47\linewidth}
\centering
\includegraphics[width=\linewidth]{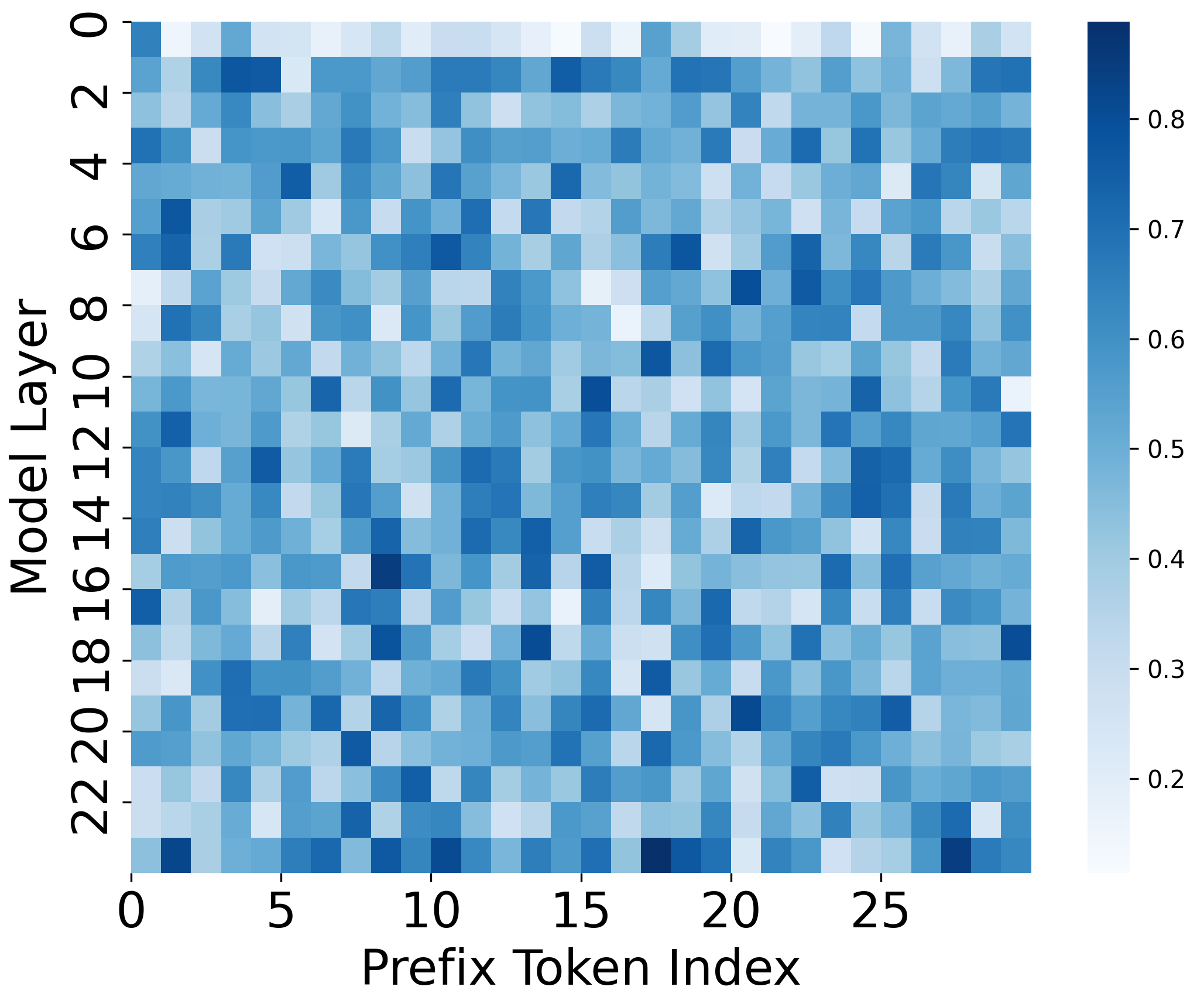}
\end{minipage}%
}%
 \caption{The weight distribution of prefix tokens, which is guided by the target event using the event-guided adaptive gating mechanism.}
    \label{prefix token weight}
\end{figure}

We demonstrate the effects of EGAG visually on the ACE05 dataset. 
Fig.~\ref{prefix token weight} shows the weight distribution of prefix tokens guided by the target event. We can observe that the target event has different attentions on different layers and different prefix tokens within the same layer. This indicates that with EGAG, the target event is adaptively capturing its relationship with other events, thus further extracting relevant information from the prefix tokens at each layer.

\section{Experimental Setup}

\subsection{Datasets}
\label{datasets}

\begin{table*}[t]
    \centering
    \resizebox{0.7\textwidth}{!}{
    \begin{tabular}{l|cccc}
    \toprule
     Dataset & ACE05 & RAMS & WIKIEVENTS & MLEE \\
     \midrule
    \# Event types & 33 & 139 & 50 & 23 \\
    \# Args per event & 1.19 & 2.33 & 1.40 & 1.29 \\
    \# Events per text & 1.35 & 1.25 & 1.78 & 3.32 \\
    \midrule
    \# Events &  & & & \\
    Train & 4202 & 7329 & 3241 & 4442 \\ 
    Dev & 450 & 924 & 345 & - \\
    Test & 403 & 871 & 365 & 2200 \\
    \bottomrule
    \end{tabular}
    }
    \caption{Statistics of datasets.}
    \label{tab_dataset}
\end{table*}

\vspace{10pt}
\noindent
\textbf{ACE05} \cite{doddington2004automatic}\footnote{https://catalog.ldc.upenn.edu/LDC2006T06
} is a joint information extraction dataset that compiles news, telephone conversations, and broadcast news, providing annotations for entities, relations, and events in English, Chinese, and Arabic. We utilize only its English event annotations for sentence-level EAE task, adopting the data preprocessing methods used by \citet{wadden-etal-2019-entity}. This dataset assumes that events are contained within a single sentence, rather than in a document, which is more typical in real-world scenarios.

\vspace{10pt}
\noindent
\textbf{RAMS} \cite{ebner-etal-2020-multi}\footnote{https://nlp.jhu.edu/rams/} is a document-level EAE dataset focused on English online news. Each example is a document consisting of five sentences, where the trigger word indicates a pre-defined event type, and their arguments are dispersed throughout the document. Therefore, a major challenge of this dataset is to address the long-distance dependencies between event triggers and arguments.

\vspace{10pt}
\noindent
\textbf{WIKIEVENTS} \cite{li-etal-2021-document}\footnote{https://github.com/raspberryice/gen-arg} is a document-level EAE dataset. The documents in this dataset are derived from English Wikipedia articles that describe real-world events, followed by the retrieval of related news articles based on reference links. They also annotate coreference links for arguments; however, in this task, we only utilize annotations for conventional arguments. Compared to RAMS, WIKIEVENTS contains a greater number of multi-event documents.

\vspace{10pt}
\noindent
\textbf{MLEE} \cite{pyysalo2012event}\footnote{
http://www.nactem.ac.uk/MLEE/
} is a document-level event extraction dataset containing manually annotated summaries of biomedical publications written in English. We follow the preprocessing procedure outlined by \citet{trieu2020deepeventmine}. As the preprocessed data only includes training/test splits, we follow \citet{he2023revisiting} to use the training set as the development set. Moreover, overlapping events are more common in MLEE than in the other three datasets.

\vspace{10pt}
\noindent
\textbf{Data Statistics.} Table~\ref{tab_dataset} shows detailed statistics for each dataset.

\subsection{Implementation details}
\label{hyperparameters settings}

\begin{table*}[t]
    \centering
    \begin{tabular}{cccccc}
    \toprule
    Hyperparameters & Search Range & ACE05 & RAMS & WIKIEVENTS & MLEE \\
    \midrule
       Batch Size  & [2, 16] & 16 & 4 & 4 & 4 \\
       Training Steps & - & 10000 & 10000 & 10000 & 10000 \\
       Window Size & - & 250 & 250 & 250 & 250 \\ 
       Max Span Length & - & 10 & 10 & 10 & 10 \\ 
       Learning Rate & - & 2e-5 & 2e-5 & 2e-5 & 2e-5 \\
       Warmup Ratio & - & 0.1 & 0.1 & 0.1 & 0.1 \\ 
       Max Gradient Norm & - & 5 & 5 & 5 & 5 \\ 
       Max Encoder Seq Length & - & 200 & 500 & 500 & 500  \\
       Max Decoder Seq Length & - & 250 &200 & 360 & 360 \\ 
       IOP Length & [10, 100] & 70 & 40 & 40 & 70 \\ 
       TOP Length & [5, 50] & 30  & 20 & 10 &  40  \\ 
       
       \bottomrule
    \end{tabular}
    \caption{Hyperparameters for DEGAP. The hyperparameters without a search range are set to the same values as in \citet{he2023revisiting}.}
    \label{hyperparameters_tab}
\end{table*}

We implement DEGAP using PyTorch and conduct experiments on a single NVIDIA GeForce RTX 3090 24G. For simplicity, we randomly initialize the embeddings of the prefix. Following TabEAE's setting \cite{he2023revisiting}, we use the first 17 layers of RoBERTa-large \cite{liu2019roberta} to instantiate the encoder, and the weights of the remaining 7 layers to initialize the weight of self-attention layers and feedforward layers of the decoder. We also randomly initialize the weight of cross-attention parts of the decoder. The optimization of our model employs the AdamW optimizer \cite{loshchilov2017decoupled} equipped with a linear learning rate scheduler. Empirical studies on the selection of the number of encoder/decoder layers can be found in \cite{he2023revisiting}.

Most hyperparameters are the same as those in TabEAE \cite{he2023revisiting}. For hyperparameters requiring tuning, we conduct grid searches in specific ranges for each dataset, choosing values that maximize the Arg-C F1 on the development set. The search ranges and the final hyperparameter values are listed in Table~\ref{hyperparameters_tab}.

\subsection{Baselines}
\label{baselines}
We compare our method with the following state-of-the-art models:
\begin{itemize}
%     \item \textbf{TSAR} \cite{xu-etal-2022-two}: A classification-based model that first identifies candidate spans and then predicts their roles within events. It utilizes an abstract meaning representation (AMR) graph to help learn the better representations of candidate spans. We report the results
% from \citet{he2023revisiting}.
%     \item \textbf{SCPRG} \cite{liu2023enhancing}: A classification-based model that enhances the final candidate span representations using context information filtered by event trigger and role interaction features. We use their code\footnote{https://github.com/LWL-cpu/SCPRG-master} to test its performance on all datasets.
    \item \textbf{Bart-Gen} \cite{li-etal-2021-document}: A conditional generation model that uses a sequence-to-sequence model and pre-defined templates to  sequentially 
    generate arguments. We report the results from \citet{he2023revisiting}.
    \item \textbf{DEGREE} \cite{hsu-etal-2022-degree}: A conditional generation model. In addition to templates, event keywords, and event descriptions are also used as auxiliary information. We report the results
from \citet{hsu2023ampere}.
% by training an AMR encoder (based on BART-large or RoBERTa-large), 
    \item \textbf{AMPERE} \citep{hsu2023ampere}: A model that builds upon DEGREE by integrating AMR information into the model using prefixes. Although comparing it directly with our method may not be fair (as it additionally trains an AMR encoder implemented by BART-large or RoBERTa-large, thus has about twice the parameters of our model), we still list the performance of its BART version for reference.
    \item \textbf{APE} \cite{zhang2023overlap}: A conditional generation model. The original paper offers Multi and Single versions trained on multiple and single datasets, respectively. For a fair comparison, we report the performance of its Single version.
    \item \textbf{AHR} \cite{ren-etal-2023-retrieve}: A model that utilizes retrieved relevant instances to facilitate sequence generation. The original paper offers three retrieval strategies: Context-Consistency Retrieval, Schema-Consistency Retrieval, and Adaptive Hybrid Retrieval. We report their best-performing method, Adaptive Hybrid Retrieval (AHR).
    \item \textbf{R-GQA} \cite{du-ji-2022-retrieval}: A model that generates arguments given a retrieved demonstration and natural questions. We report the results from the original paper.
    \item \textbf{PAIE} \cite{ma2022prompt}: A QA-based model that extracts arguments by querying context using role slots in the event template. We report the results
from \citet{he2023revisiting}.
    \item \textbf{SPEAE} \cite{nguyen2023contextualized}: A model based on PAIE that utilizes prefixes to integrate relevant context information into the model and alleviate sub-optimal manual prompts. We report the results
from the original paper.
    \item \textbf{TabEAE} \cite{he2023revisiting}: Building upon PAIE, it simultaneously extracts multiple events and replaces the base model BART used by PAIE with RoBERTa. The original paper introduces three training-inference paradigms: Multi-Multi, Multi-Single, and Single-Single (consistent with our training-inference paradigm). To ensure a fair comparison and visually demonstrate the effectiveness of our proposed prefix, we report the performance of its Single-Single version.
    % "The Multi-Single and Multi-Multi versions of TabEAE demonstrate SOTA performance only on certain datasets (while significantly underperforming our method on others), respectively, making it difficult to choose one version for comparison. Additionally, the goal of this paper is to demonstrate the effectiveness of the proposed prefix. Therefore, the Single-Single version was selected."
\end{itemize}

\subsection{The implementation details of DRAAE}
\label{The implementation details of DRAAE}

We follow \cite{ren-etal-2023-retrieve} to use a \textbf{frozen} S-BERT \cite{reimers2019sentence} to retrieve the top-$k_{ins}$ relevant instance $\textbf{d}_{ins}$ and the top-$k_{tem}$ relevant templates $\textbf{d}_{tem}$ from training corpus for the input instance and template, respectively. Then, we use RoBERTa-large as an encoder to obtain the retrieved instance embedding $\textbf{h}_{\textbf{d}_{ins}}$ and template embedding $\textbf{h}_{\textbf{d}_{tem}}$ through the representation of the first input token:
\begin{equation*}
    \textbf{h}_{\textbf{d}_{ins}} = \text{RoBERTa}(\textbf{d}_{ins}),
\end{equation*}
\begin{equation}
    \textbf{h}_{\textbf{d}_{tem}} = \text{RoBERTa}(\textbf{d}_{tem}),
    \label{eq15}
\end{equation}
where $\textbf{h}_{\textbf{d}_{ins}} = \{\textbf{h}_{\textbf{d}_{ins}}^{(1)}, 
 \textbf{h}_{\textbf{d}_{ins}}^{(2)}, ...,  
  \textbf{h}_{\textbf{d}_{ins}}^{(k_{ins}) }  \}$, and $\textbf{h}_{\textbf{d}_{tem}} = \{\textbf{h}_{\textbf{d}_{tem}}^{(1)}, 
 \textbf{h}_{\textbf{d}_{tem}}^{(2)}, ...,  
\textbf{h}_{\textbf{d}_{tem}}^{(k_{tem}) }  \}$. These retrieved instance contexts and event templates are incorporated into each transformer layer's self-attention module. 

The new self-attention with retrieved event information is formulated as:
\begin{equation*}
H \leftarrow \text{LayerNorm}(H^{'} + H),
\end{equation*}
\begin{equation}
H^{'} = \text{MHSA}(\textbf{h}_{\textbf{d}} \oplus H)_{|\textbf{h}_{\textbf{d}}|:|\textbf{h}_{\textbf{d}} \oplus H|},
\end{equation}
where $\textbf{h}_{\textbf{d}}$ represents  either $\textbf{h}_{\textbf{d}_{ins}}$ or $\textbf{h}_{\textbf{d}_{tem}}$, $H$ is the hidden state. 

Finally, we use RoBERTa in Eq.~\ref{eq15} to encode the input context and event template in the same manner as in DEGAP (\S~\ref{Encoding with Prefixes}), and further extract argument spans (\S~\ref{Span Selection}). 

\section{Experimental Analysis}

\subsection{Performance Comparison with LLMs}
\label{Performance Comparison with LLMs}

In this section, we compare our method with three publicly available LLMs: GPT-3 \cite{brown2020language} in its text-davinci-003 and GPT-3.5-turbo-instruct versions \cite{ouyang2022training}, and GPT-4 \cite{achiam2023gpt}.

\begin{table}[ht]
    \centering
    \resizebox{1.0\columnwidth}{!}{
    \begin{tabular}{l|cc}
    \toprule
    \multirow{2}{*}{Model} & \multicolumn{2}{c}{RAMS} \\
     & Arg-I & Arg-C \\   \midrule
     text-davinci-003 \cite{brown2020language} &  46.2 & 39.6 \\
     GPT-3.5-turbo-instruct \cite{ouyang2022training} & 43.3 & 37.0 \\
     GPT-4 \cite{achiam2023gpt} & 52.4 & 44.1 \\
       \midrule
     DEGAP (Ours)  & \textbf{58.5} & \textbf{54.2} \\
     \bottomrule
    \end{tabular} 
    }
    \caption{Performance comparison with LLMs. We report their performance from \cite{zhou2023heuristics}.}
    \label{tab_performance compare with llms}
\end{table}

Table~\ref{tab_performance compare with llms} shows that our method achieves improvements of 6.1\% and 10.1\% in Arg-I and Arg-C F1 scores respectively, emphasizing our method's significant advantage over LLMs. This makes sense because our model are trained using task-specific data, whereas LLMs can only perform in-context learning through one or two demonstrations.  
This observation is also consistent with previous findings, suggesting that despite LLMs making substantial progress in many natural language processing tasks \cite{zhao2023survey, muennighoff2022crosslingual} such as text generation tasks \cite{qin2023chatgpt}, further efforts are still needed for them to excel in generating complex structured outputs (e.g., generating structured event information) \cite{tang2023struc,chen2024large}. 

\subsection{Efficiency Analysis}
\label{Efficiency Analysis}

\begin{table}[ht]
    \centering
    \resizebox{1.0\columnwidth}{!}{
    \begin{tabular}{l|cccc}
    \toprule
    Model & ACE05 & RAMS & WIKIEVENTS & MLEE \\
    \midrule
        DRAAE &  61.18 &   256.87 & 121.76  & 626.31  \\ 
        TabEAE & 3.69 & 41.35 & 18.11 & 71.96 \\
        DEGAP & 4.36  &    43.36     &   19.64 &  88.60 \\
    \bottomrule
    \end{tabular}
    }
    \caption{Inference time (second) for different models on test set of four datasets. Experiments are run on one same  NVIDIA GeForce RTX 3090.}
    \label{efficiency analysis}
\end{table}

Table~\ref{efficiency analysis} reports the overall inference time for DEGAP (w/o retrieval) and DRAAE (w/ retrieval). It can be observed that DEGAP runs approximately 6-7 times faster than DRAAE on RAMS, WIKIEVENTS, and MLEE, and approximately 15 times faster than DRAAE on ACE05. This is mainly because: 1) DRAAE requires additional time costs to retrieve the top-$k$ similar event instances and templates; 2) For the retrieved event instances and templates, DRAAE needs to feed them into the encoder to obtain meaningful embeddings. In contrast, DEGAP does not require retrieval and further encoding. Only equipped with a set of learnable prefixes,  our DEGAP can adaptively learn semantic information from different event instances and templates during training, and provide relevant guidance at each model layer during inference.  Note that due to the parallelized attention calculation for the introduced prefixes and original input on the GPU, there is a negligible impact on inference speed compared to the base model TabEAE.

\subsection{Prefix length}
\label{Prefix length}

\begin{figure}[ht]
\centering
\subfigure{
\begin{minipage}[t]{0.49\linewidth}
\centering
\includegraphics[width=\linewidth]{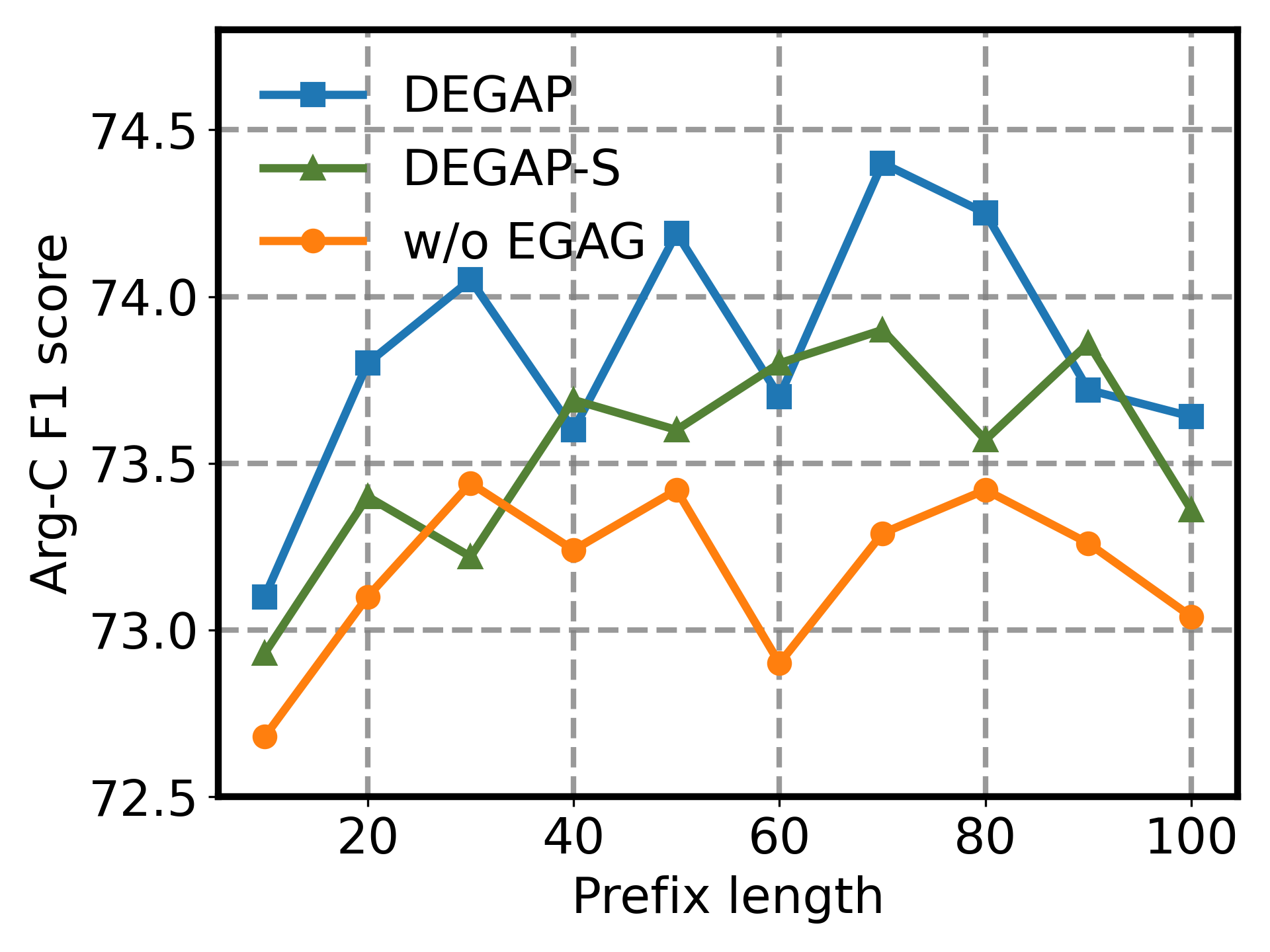}
\end{minipage}%
}%
\subfigure{
\begin{minipage}[t]{0.49\linewidth}
\centering
\includegraphics[width=\linewidth]{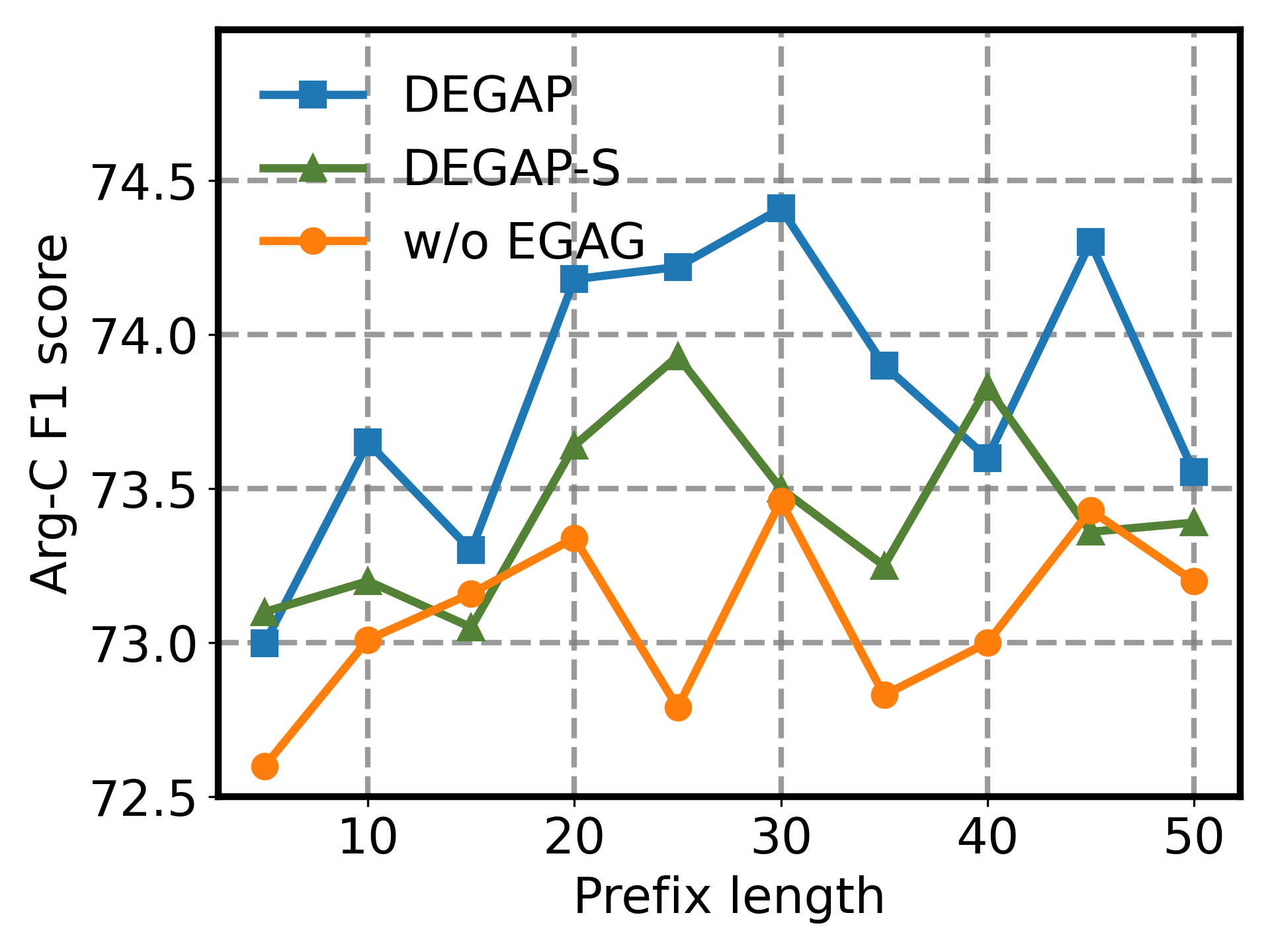}
\end{minipage}%
}%
\caption{Results under different prefix lengths.}
    \label{prefix len}
\end{figure}

We further analyze the impact of prefix length on model performance. Intuitively, a longer prefix implies more trainable parameters, thus enabling it to provide more expressive guidance for the model regarding relevant semantic information. As shown in Fig.~\ref{prefix len}, as the prefix length increases, the model performance tends to improve, followed by a slight decline. We speculate that this is because too long prefixes provide redundant semantic information, making it difficult for the model to focus on extracting the target event.

\subsection{Event Template Variant}

As described in \S~\ref{Event Template Design}, the event template $\mathcal{T}_{e}$ consists of two parts: the type part $\mathcal{I}_{e}$ and the schema part $\mathcal{S}_{e}$. To explore the necessity of type part $\mathcal{I}_{e}$, following \citet{shi2023hybrid}, we use type markers “\texttt{<Event type>}” and “\texttt{</Event type>}” as learnable special tokens, inserting them respectively into the corresponding schema part $\mathcal{S}_{e}$ as the template for event type $e$. Additionally, we compute Eq.~\ref{a_tem} by averaging the representations of the type markers. 

\begin{table}[ht]
    \centering
    \resizebox{1.0\columnwidth}{!}{
    \begin{tabular}{l|cccc}
    \toprule
       Model  & ACE05 & RAMS & WIKIEVENTS & MLEE \\  \midrule
        w/ type part &  \textbf{74.4}	& \textbf{54.2} &	67.1	& \textbf{73.4} \\ 
        r/ type part	& 74.2	& 53.9	& \textbf{67.3}	& 73.1  \\
        \bottomrule
    \end{tabular}
    }
    \caption{Analysis of the necessity for the Type part. The numbers in the table all refer to the Arg-C F1 scores. \textbf{r/}: Replace the type part with type markers.}
    \label{tab_analysis of type part}
\end{table}

The experimental results in Table~\ref{tab_analysis of type part} suggest that describing the event type with natural language sentence generally outperforms the use of type markers. Consequently, we retain this setting in our event template $\mathcal{T}_{e}$. Note that as our method does not rely on type markers that need to be learned, it can more easily adapt to new event types.

\end{document}